\documentclass{article}
\usepackage{ijcai25}

\usepackage{times}
\usepackage{soul}
\usepackage{url}
\usepackage[hidelinks]{hyperref}
\usepackage[utf8]{inputenc}
\usepackage[small]{caption}
\usepackage{graphicx}
\usepackage{amsmath}
\usepackage{amsthm}
\usepackage{booktabs}
\usepackage{algorithm}
\usepackage{algorithmic}
\usepackage[switch]{lineno}

\usepackage{multirow}
\usepackage{subfigure}

\title{A Dynamic Knowledge Update-Driven Model with Large Language Models for Fake News Detection}

\author{
Di Jin$^{1,2}$\and
Jun Yang$^1$\and
Xiaobao Wang$^{1,}$\thanks{Corresponding Author}\and
Junwei Zhang$^3$\and
Shuqi Li$^1$\And
Dongxiao He$^1$\\
\affiliations
$^1$Tianjin Key Laboratory of Cognitive Computing and Application, College of Intelligence and Computing, Tianjin University, Tianjin, China\\
$^2$Key Laboratory of Artificial Intelligence Application Technology, Qinghai Minzu University,
China\\
$^3$Hangzhou Institute of Medicine, Chinese Academy of Sciences, Hangzhou, China\\
\emails
\{jindi, yangjun, wangxiaobao, junwei, shuqili, hedongxiao\}@tju.edu.cn
}

\begin{document}

\maketitle

\begin{abstract}
    As the Internet and social media evolve rapidly, distinguishing credible news from a vast amount of complex information poses a significant challenge. Due to the suddenness and instability of news events, the authenticity labels of news can potentially shift as events develop, making it crucial for fake news detection to obtain the latest event updates. Existing methods employ retrieval-augmented generation to fill knowledge gaps, but they suffer from issues such as insufficient credibility of retrieved content and interference from noisy information. We propose a \textbf{DY}namic k\textbf{N}owledge upd\textbf{A}te-driven \textbf{MO}del for fake news detection (\textbf{DYNAMO}), which leverages knowledge graphs to achieve continuous updating of new knowledge and integrates with large language models to fulfill dual functions: news authenticity detection and verification of new knowledge correctness, solving the two key problems of ensuring the authenticity of new knowledge and deeply mining news semantics. Specifically, we first construct a news-domain-specific knowledge graph. Then, we use Monte Carlo Tree Search to decompose complex news and verify them step by step. Finally, we extract and update new knowledge from verified real news texts and reasoning paths. Experimental results demonstrate that DYNAMO achieves the best performance on two real-world datasets.
\end{abstract}

\section{Introduction}
With the rapid development of the Internet and social media, fake news has been proliferating amid the high-speed dissemination of information~\cite{Backdoor2025Jin,Elevating2025Wang,Commonsense2023Yu,General2024Zhu}. Misleading information, false advertisements, fabricated accounts, and other statements all seriously impact the online experience of users and the security of the network environment~\cite{Fake2017Shu,Generalized2023Dong,Augmenting2023Wang}. Given the complexity and diversity of information types, identifying credible news among a vast amount of information is challenging~\cite{Multi2024Ma}. Compared to human judgment, automated technical means for determining the authenticity of news are better suited to handle vast and complex information on the Internet. 

Traditional methods for detecting fake news can be categorized into three types based on the information required: content-based methods~\cite{Compare2021Hu,Capturing2020Przybyla,Unveiling2024Dong,Multimodal2024Zhu}, user-based methods~\cite{Proactive2020Chen,User2021Dou} and propagation-based methods~\cite{Rumor2020Bian,Jointly2023Sun}. These approaches leverage various types of news-related information and have achieved good results in detection accuracy. However, due to their black-box nature~\cite{Explainable2024Wang}, the models cannot provide explicit reasons to enhance the credibility of detection results. Additionally, traditional methods often require a large amount of manually annotated data for model training. High-quality annotated samples are not only time-consuming and labor-intensive to obtain, but also require annotators to have sufficient understanding of events to assign accurate labels~\cite{Large2024Tan}, making data collection a pressing issue.

To address the data issue in fake news detection, some studies have begun to consider unsupervised detection methods. Large language models (LLMs), renowned for their remarkable zero-shot learning and advanced reasoning abilities, have been introduced into fake news detection. Using techniques such as chain of thought to guide LLMs in outputting their thinking and reasoning processes~\cite{Explainable2024Wang,Fact2023Pan}, reasonable and informative rationales can be provided from multiple perspectives~\cite{Bad2024Hu}, effectively enhancing interpretability. 
However, LLMs are insensitive to real-world dynamics~\cite{Think2024Sun}. News events are often sudden and unstable, and their development can lead to shifts in the authenticity of news. Therefore, promptly updating the latest developments of events and acquiring the newest information are crucial for fake news detection.

The substantial update costs associated with LLMs hinder the timely updating of their internal knowledge, making it difficult to meet the high timeliness demands of fake news detection~\cite{MindMap2024Wen}. 
A direct and effective approach to address the timeliness issue is to use Retrieval-Augmented Generation (RAG). RAG retrieves information from external resources, such as web pages, in an effort to capture the latest event developments~\cite{Re2024Li,VeraCT2024Niu}. However, RAG is not without its flaws. It faces challenges such as insufficient credibility of the retrieved content, interference from noisy information, retrieval inefficiency caused by information overload, and network latency issues~\cite{Benchmarking2024Chen}, which may lead to hallucination issues in LLMs and reduced efficiency in fake news detection~\cite{Evaluating2024Adlakha}. 

\begin{figure}
    \centering
    \includegraphics[width=8.6 cm]{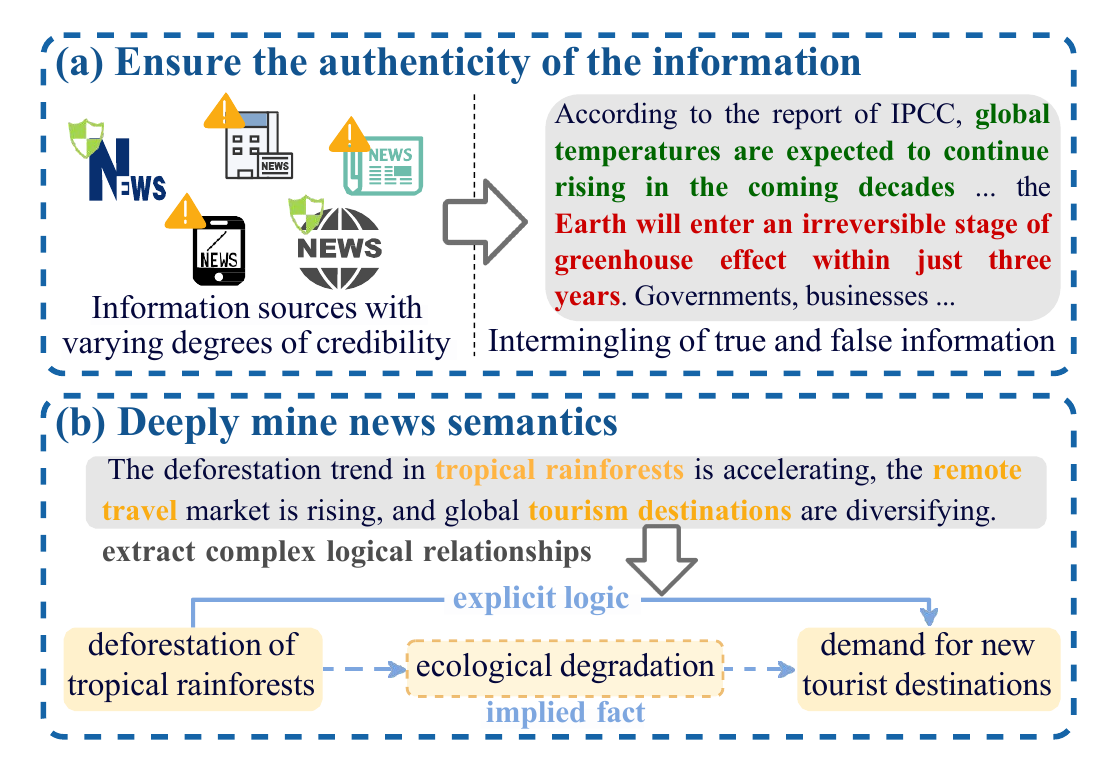}
    \caption{Two key challenges of knowledge updating. (a) Ensuring the authenticity of the information: The varying reliability of information sources and the mixture of true and false information make verifying authenticity challenging. (b) Deeply mining news semantics: The demand for concise news expression results in complex semantics and multi-layered logical relationships being concealed. }
    \label{fig:Motivation}
    \centering
\end{figure}

Given the significant advantages of knowledge graphs in adaptability and expansibility, maintenance costs, knowledge reliability, and retrieval efficiency~\cite{Unifying2024Mou,Knowledgeable2024Zhang}, we explore the possibility of using knowledge graphs to address the timeliness issue in fake news detection, envisioning the continuous integration of new information (both labeled and unlabeled) into the knowledge graph. Furthermore, due to the timeliness of news, which allows for prompt reporting on the latest events, we choose news as the source of new knowledge. However, as shown in Figure~\ref{fig:Motivation}, implementing this idea requires overcoming two key challenges: (1) Ensuring the authenticity of the information: Due to the complexity of knowledge sources and the intermingling of true and false information, verifying the authenticity of the information when updating the knowledge graph is an extremely challenging task. (2) Deeply mining news semantics: News is often rich in complex semantic information and multi-layered logical relationships. How to accurately and comprehensively mine the rich information embedded in news presents another major challenge. 

Addressing the aforementioned issues, we propose a \textbf{DY}namic k\textbf{N}owledge upd\textbf{A}te-driven \textbf{MO}del for fake news detection (\textbf{DYNAMO}). The model simultaneously performs fake news detection and verifies the accuracy of new knowledge. It achieves dynamic knowledge updates in fake news detection by updating the knowledge graph, and continuously injecting new knowledge to enhance LLMs' understanding of various news events. This improves the model's ability to adapt to event developments, effectively addressing the timeliness requirements of fake news detection. Specifically, we first utilize the LLM to assist in constructing a news-domain-specific knowledge graph as external supplemental knowledge. Next, the knowledge graph aids the LLM in detecting fake news. Using Monte Carlo Tree Search, the news is broken down into multiple sub-questions, which are answered sequentially to verify the authenticity of the news. Additionally, we predict the authenticity of new knowledge by examining the news content. Through Monte Carlo Tree Search, the model explores different news splitting combinations to identify the optimal splitting approach, thus unveiling the deep logical relationships within the news, which is beneficial for subsequent new knowledge extraction. Besides, by splitting the news, the model can gradually verify the authenticity of the content, reducing the chances of missing information points and improving detection accuracy and credibility. Finally, for news that is verified as true, the LLM extracts knowledge triples from the news text and reasoning paths, enabling efficient and convenient knowledge updates.

To summarize, the contributions of this paper include:
\begin{itemize}
    \item We identify a new issue that needs attention in fake news detection. Due to the development of news events, which may result in alterations to news labels, there is a timeliness requirement for news detection.
    \item We propose a dynamic knowledge update-driven model that innovatively verifies the correctness of new knowledge through detecting the authenticity of news, while simultaneously extracting knowledge from real news. 
    \item Extensive experiments on two real-world datasets demonstrate that the proposed model achieves the state-of-the-art performance on various metrics.
\end{itemize}

\section{Related Work}

\subsection{Traditional Methods in Fake News Detection}
Content-based methods determine the authenticity of news by analyzing features such as text, topic selection, and writing style. ~\cite{Capturing2020Przybyla} focuses on typical words in fake news to learn the emotional language elements of news for detection. As for news spreading on social media, additional social context information can serve as supplementary information to assist in detecting fake news. 
User-based methods believe that factors such as users' social behaviors and attitudes can reveal the likelihood of users spreading fake news. UPFD~\cite{User2021Dou} obtains user preferences by analyzing users' historical posts on social media. 
Propagation-based methods detect fake news by extracting features of the propagation network. Bi-GCN~\cite{Rumor2020Bian} considers both the ‘‘top-down" propagation and the ‘‘bottom-up" dissemination structure to learn the causal and structural characteristics of rumor propagation. 
However, due to their black-box nature, the models lack interpretability. Furthermore, traditional models in fake news detection rely on high-quality annotated data for model training, which is difficult to collect. 

\begin{figure*}
    \centering
    \includegraphics[width=15.2 cm]{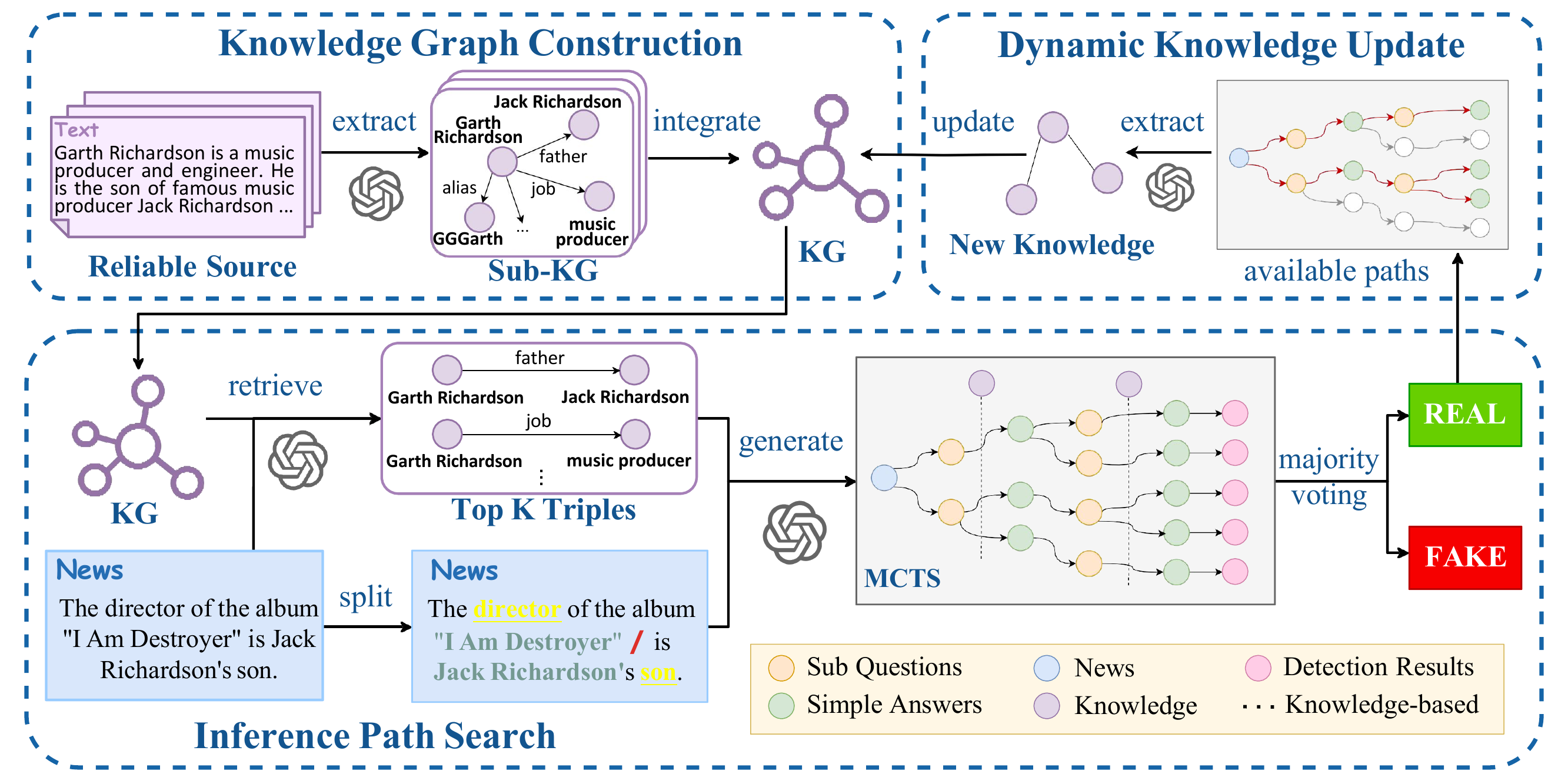}
    \caption{The overview of our proposed model DYNAMO.}
    \label{fig:Overview}
    \centering
\end{figure*}

\subsection{LLMs in Fake News Detection}
To alleviate the issues of data collection and interpretability, some approaches have introduced Large Language Models (LLMs) as detectors. ProgramFC~\cite{Fact2023Pan} decomposes complex claims into simpler subtasks and processes them sequentially, guiding the LLM to provide explicit reasoning processes. L-Defense~\cite{Explainable2024Wang} utilizes the reasoning abilities of the LLM to conduct reasoning and judgment on two opposing sets of evidence, and generate a summary of reasons.
Due to the lack of domain-specific expertise in news field by LLMs, some approaches introduce knowledge graphs to provide external knowledge for assistance. PEG~\cite{Unifying2024Mou} refines external knowledge through semantic filtering and aggregates it into global knowledge. DALK~\cite{Unifying2024Mou} collects evidential subgraphs through path-based and neighbor-based exploration. To address the timeliness requirements of fake news detection, some approaches adopt Retrieval-Augmented Generation. STEEL~\cite{Re2024Li} and VeraCT Scan~\cite{VeraCT2024Niu} retrieve and extract key information from web as external knowledge, thereby mitigating the issue of outdated knowledge in LLMs. However, due to lack of verification of the retrieved information, it may instead mislead the LLM into making incorrect judgments. Furthermore, it fails to truly achieve knowledge updating.

\section{Methods}

Figure \ref{fig:Overview} presents the overview of our model. Firstly, we extract knowledge from news-related texts generated by reliable sources to construct a news-domain-specific knowledge graph. Secondly, we utilize the knowledge graph to assist the LLM in completing fake news detection. Lastly, we leverage the LLM to extract new knowledge from authentic news and reasoning paths for updating the knowledge graph.

\subsection{Task Definition}

There is a news-domain-specific knowledge graph $G = (E, R, S)$, where $E$ is the set of entities, $R$ is the set of relations and $S \subseteq E \times R \times E$ is the set of triples. $L$ denotes the LLM used for detection. The task of our model is to predict the label $y \in \{\text{Real}, \text{Fake}\}$ for a given news article $c$ involving the knowledge graph.
Mathematically, $y = L(c,G)$.

\subsection{Knowledge Graph Construction}
\label{sec:KGC}

Knowledge graphs possess structured formats and expansibility~\cite{DALK2024Li}, hence we select knowledge graphs as the carrier for knowledge retrieval and updating. 
The existing knowledge graphs lack the relevant knowledge to support fake news detection.
To enhance LLMs' understanding of news, we first construct a news-domain-specific knowledge graph $G$, integrating various types of information involved in news events in a structured manner. The source of knowledge is news-related texts generated by reliable sources.

\textbf{Entity-Centered Knowledge Extraction.} 
Firstly, key entities are extracted from the text to form an entity set $E$. To obtain more knowledge-rich triples, inspired by~\cite{DALK2024Li}, we adopt a generative approach to extract relations between entities. Thus, the vocabulary used to describe relations between entities is not limited by the original text. The entity set and text are input into the LLM, which is instructed to extract keywords for describing the relations between relevant entities or directly generate new summary terms $R$, and output the corresponding knowledge triples $S$. 

\textbf{Event-Centered Knowledge Extraction.} 
Furthermore, due to the inclusion of multifaceted factors such as time and location in facts, these elements cannot be simply summarized as two entities and the relationship between them. Consequently, we also adopt the form of (subject, predicate, object) to guide the LLM in extracting triples for events. 

Thus, we obtain a comprehensive news-domain-specific knowledge graph and it is used to assist subsequent modules.

\subsection{Inference Path Search}

Due to the pursuit of concise expression in news reporting, many complex semantics and logical relationships may be summarized in brief statements, leading to increased difficulty in understanding. We break down complex news into several simple sub-questions, answer and verify their truthfulness sequentially, thereby dissecting the underlying semantic relationships and reducing the difficulty for the LLM to understand and judge. Additionally, breaking down the news allows the LLM to output its thinking and reasoning processes, enhancing interpretability. At the same time, by requiring the LLM to output intermediate results, we prompt it to more fully utilize its internal knowledge and provide richer information for subsequent updates to the knowledge graph.

\textbf{Monte Carlo Tree Search.} 
Inspired by~\cite{Mutual2024Qi}, we utilize Monte Carlo Tree Search (MCTS) with multiple types of actions to generate reasoning paths, thereby accomplishing the breakdown and verification of news. MCTS is capable of finding approximately optimal strategies within limited time, thus it is ideal for large-scale problems. Additionally, the algorithmic structure of MCTS is relatively simple, facilitating easy implementation and extension.

For each piece of news $c$, the model constructs a search tree $T$. The root node of $T$ represents the news $c$, an edge represents an action $a$, and each child node is an intermediate step $s$ generated by the LLM $L$ based on the corresponding action. The path from the root node to a leaf node $s_d$ constitutes a candidate reasoning path $t = (c, s_1, s_2, \ldots, s_d)$, where $d \leq h$ and $h$ is the pre-defined height limit of the tree. From the search tree $T$, we can extract a set of reasoning paths $T_{\text{route}} = \{t_1, t_2, \ldots, t_n\}$, where $n \geq 1$ and $n$ is the number of searches conducted.

To enhance detection efficiency and adapt the model to fake news detection tasks, we define three types of actions:

$A_1$: Generate a sub-question. In this step, the LLM is required to propose the next sub-question to be answered based on the previous question-and-answer process, thereby breaking down the news into simpler content that is easier for the LLM to verify as true or false.

$A_2$: Answer a sub-question. In this step, the LLM utilizes its own knowledge to answer the sub-question generated in step $A_1$, thereby prompting the LLM to generate a detailed and reliable reasoning process and providing more knowledge for subsequent updates to the knowledge graph.

$A_3$: Provide a detection result. If the current question-and-answer process generated contains sufficient information to determine the authenticity of the news, the LLM is allowed to directly give the final prediction result in this step.

Next, we introduce how to select actions. For explored nodes, reward scores are assigned to the nodes during the back-propagation process. Initially, the reward score for all nodes is defined as $Q(s) = 0$. We define the reward score based on the current reasoning results as follows:
\begin{equation}
    r = \frac{p_{\text{major}}}{p_{\text{major}} + p_{\text{minor}}},
\end{equation}
where $p_{\text{major}}$ is the number of reasoning paths whose results belong to the majority outcome, and $p_{\text{minor}}$ is the number of reasoning paths with minority outcomes.

When a leaf node $s_d$ appears, if the judgment result of this leaf node $s_d$ is consistent with the judgment results of the current majority, then a reward will be assigned to each node on the corresponding reasoning path $t'$ of $s_d$, that is:
\begin{equation}
    Q(s_i) = Q(s_i) + r, \quad s_i \in t'.
\end{equation}

Then, the Upper Confidence Bounds applied to Trees (UCT) strategy is used for node selection, with the specific formula as follows:
\begin{equation}
\label{equ:uct}
    UCT(s_j) = \frac{Q(s_j)}{V(s_j)} + \alpha\sqrt{\frac{\ln V_{\text{parent}}(s_j)}{V(s_j)}}, \quad s_j \in T.
\end{equation}
Here, \(Q\) represents the node score, \(\alpha\) is a constant, \(V\) denotes the visit count, and \(V_{\text{parent}}\) is the visit count of the parent node.

At each timestamp \(i\), MCTS randomly selects an unexplored node \(s_i\) and uses a prompt to guide the LLM to generate the relevant content for \(s_i\) based on the current path \((c, s_1, s_2, \ldots, s_{i-1})\), or selects an explored node according to the UCT strategy. Additionally, there exists a precedence relationship among different action types, which is manifested as follows: $A_2$ can only occur after $A_1$, and $A_3$ can only occur after the root node and $A_2$. For the final detection result of news $c$, we adopt the majority voting, where the majority result among multiple leaf nodes is selected as the final result.

\textbf{Knowledge Graph Enhancement.}
To enhance the understanding of news by the LLM, we utilize the knowledge graph constructed in Section ~\ref{sec:KGC} as the external knowledge to assist in detection. During Action $A_2$ of the MCTS, news-related knowledge triples can be provided as supplementary knowledge to the LLM, thereby augmenting its capabilities. Firstly, we extract key entities from the sub-question to form an entity set for the question, denoted as $E_Q$. We then match the entities in $E_Q$ with those in the knowledge graph $G$ to obtain a matched entity set $E_G$. From $E_G$, we construct a problem-specific knowledge subgraph $G_Q$ comprising the one-hop triples associated with the entities in $E_G$. Due to the limited input text length of LLMs, we filter the retrieved triples. We use a prompt to guide the LLM in ranking the knowledge triples based on their relevance to the sub-question, and select the top-$K$ triples as the final supplementary knowledge. 

\begin{table*}
\renewcommand\tabcolsep{9.5 pt}
\centering
\begin{tabular}{cccccccccc}
\toprule
\multirow{2}{*}{Methods}  & \multicolumn{4}{c}{\textbf{Hover}} & \multicolumn{4}{c}{\textbf{Feverous}} \\
\cline{2-5} \cline{6-9} 
 & \textbf{Accuracy} & \textbf{F1} & \textbf{Precision} & \textbf{Recall} & \textbf{Accuracy} & \textbf{F1} & \textbf{Precision} & \textbf{Recall} \\
 
\midrule

RoBERTa-NLI & 0.4395 & 0.3141 & 0.6145 & 0.2109 & 0.4301 & 0.1986 & 0.5266 & 0.1224 \\
BERT-FC & 0.5858 & 0.5619 & 0.5660 & 0.5579 & 0.5548 & 0.3081 & 0.4603 & 0.2315 \\
GET & 0.5378 & 0.4398 & 0.4830 & 0.4928 & 0.5523 & 0.4427  & 0.5579 & 0.5005 \\
\midrule

ProgramFC & 0.4701 & 0.3640 & 0.7280 & 0.2427 & 0.5433 & 0.4869 & 0.7345 & 0.3641 \\
TOG & 0.5548 & 0.5915 & 0.6736 & 0.5272 & 0.6093 & 0.6599 & 0.7143 & 0.6132 \\
CoRAG & 0.5440 & 0.3824 & 0.6474 & 0.2732 & 0.6030 & 0.5912  & 0.6293 & 0.5574 \\
Phi3-mini & 0.5850 & 0.5951 & 0.7176 & 0.5083 & 0.6090 & 0.5412 & 0.7169 & 0.4347 \\
\textbf{DYNAMO} & \textbf{0.6250} & \textbf{0.7401} & \textbf{0.7367} & \textbf{0.7436} & \textbf{0.6167} & \textbf{0.7433} & \textbf{0.7384} & \textbf{0.7483} \\
\bottomrule
\end{tabular}
\caption{Performance comparison on two datasets of our
model and baselines. The best performance is in bold.}
\label{tab:performance}
\end{table*}

\subsection{Dynamic Knowledge Update}

We aim to fully leverage the knowledge acquired during the detection process to facilitate detection of similar news in the future. Additionally, dynamically updating the model's knowledge allows it to stay informed about the development of events, thereby enhancing its capability to handle real-world scenarios. Therefore, for news confirmed as true in the detection results, we extract new knowledge from the question-and-answer process involving the news text and reasoning paths, which is used to update the knowledge graph.

The method for extracting triples here is the same as the approach used in Section ~\ref{sec:KGC} for constructing the knowledge graph. Firstly, we extract key entities from the text to form a new entity set $E_\text{new}$. We then input the entity set and the text into the LLM, instructing it to extract keywords that describe the relationships between relevant entities or directly generate new summary terms $R_\text{new}$, and output the corresponding triples $S_\text{new}$. Finally, we store the newly acquired triples $G_\text{new} = (E_\text{new}, R_\text{new}, S_\text{new})$ into the knowledge graph $G$.
The updated $G$ will be utilized in the subsequent news detection, and it will continuously be updated with new knowledge extracted from news that will be verified as true.

\section{Experiments}

We first introduce the datasets, baselines, and various settings used in the experiments (Section~\ref{sec:set}). We validate the model's performance on two real-world datasets (Section~\ref{sec:result}). Additionally, we design experiments to demonstrate the advantages of dynamic knowledge updating (Section~\ref{sec:kgrenew}). Then, we conduct ablation experiments to verify the necessity of each component and the rationality of the parameter (Section~\ref{sec:ablation}). Finally, we present a case to showcase how dynamic knowledge updating aids fake news detection (Section~\ref{sec:casestudy}).

\subsection{Experiment Setting}
\label{sec:set}

\textbf{Datasets.}
To validate the enhancement effect of the dynamically updated knowledge graph, we require datasets that comprise a series of news with similar content but differing keywords and labels. Additionally, we desire the datasets to include factual statements related to the news, which can be utilized to construct a news-specific knowledge graph. In summary, two real-world datasets are selected for experiments. Hover~\cite{Hover2020Jiang} is a dataset of multi-hop factual statements, containing 18,171 samples. It extracts facts from several Wikipedia articles related to factual statements and categorizes whether the statements are correct. Feverous~\cite{Feverous2021Aly}, with 26,928 samples, extracts structured evidence, unstructured evidence, or a combination of both from Wikipedia to verify statements. Since we focus solely on data in natural language form, we follow the setup of ~\cite{Fact2023Pan} and select only sentence evidence and corresponding statements as our samples.

\textbf{Baselines.}
We compare the proposed model with the following baseline models: (1) Traditional methods: RoBERTa-NLI~\cite{Adversarial2020Nie} utilizes a finely tuned version of RoBERTa-large across four natural language inference datasets and BERT-FC~\cite{BERT2020Soleimani} trains BERT to complete fact-checking tasks. GET~\cite{Evidence2022Xu} implements graph-based semantic structure mining to learn long-distance semantic dependency relationships. (2) LLM-based methods: ProgramFC~\cite{Fact2023Pan} breaks down complex claims into simpler sub-claims and TOG~\cite{Think2024Sun} utilizes a LLM agent iteratively to perform beam search on the KG to discover the most promising reasoning paths. CoRAG~\cite{RAGAR2024Khaliq} is a real-time retrieval-based model. Phi3-mini~\cite{Phi2024Abdin} is the LLM backbone that we utilize for DYNAMO and the LLM-based baselines. Here we use the form of direct questioning to prompt Phi3-mini to assess the authenticity of the news.

\begin{table*}
\renewcommand\tabcolsep{12.5 pt}
\centering
\begin{tabular}{cccccc}
\toprule
Settings   & \textbf{Hover1}  & \textbf{Hover2} & \textbf{Hover2+kg1} & \textbf{Hover3} & \textbf{Hover3+kg1\&2}  \\
\midrule
DYNAMO &0.6309  &0.6267	&0.6400 	&0.6167	 &0.6300  \\

\bottomrule
\end{tabular}
\caption{Performance gain achieved by knowledge graph updates on Hover. The evaluation metric is accuracy in the table.}
\label{tab:gainhover}
\end{table*}

\begin{table*}
\renewcommand\tabcolsep{8 pt}
\centering
\begin{tabular}{cccccc}
\toprule
Settings & \textbf{Feverous1} & \textbf{Feverous2} & \textbf{Feverous2+kg1} & \textbf{Feverous3} & \textbf{Feverous3+kg1\&2}  \\
\midrule
DYNAMO &0.6373  & 0.6127  &	0.6195	&0.6000	& 0.6075 \\

\bottomrule
\end{tabular}
\caption{Performance gain achieved by knowledge graph updates on Feverous. The evaluation metric is accuracy in the table.}
\label{tab:gainfever}
\end{table*}

\textbf{Implementation Details.}
We select Phi3-mini-128k as the LLM backbone for DYNAMO and the LLM-based baselines. We follow the experimental setup for few-shot learning, so we set the number of samples for training and few-shot learning to 3, and adopt 5 epochs for training. For the proposed model, the number of times MCTS is executed, $n$, is 5, with the height limit $h$ of the search tree set to 5 and 9 respectively for the Hover and Feverous datasets. Each node generating 2 child nodes when a new action is executed. For knowledge graph retrieval, we select the top-5 knowledge triples as external knowledge input to the large language model. The constant $\alpha$ in the formula~\ref{equ:uct} is set to 2.

\textbf{Experimental Environment.} 
Experiments are conducted on the PyTorch 2.5.0 platform with an Intel(R) Xeon(R) Gold 6326 CPU (2.90GHz) and an NVIDIA A40 48GB GPU. 

\subsection{Experiment Results}
\label{sec:result}
Table~\ref{tab:performance} presents the performance of the proposed model and baseline models on two datasets. Relevant knowledge triples are retrieved from the knowledge graph constructed in Section~\ref{sec:KGC} to serve as external knowledge. As can be seen from Table~\ref{tab:performance}, we can draw the following conclusions:
(1) DYNAMO has achieved state-of-the-art accuracy on
two datasets. This is attributed to the Monte Carlo Tree Search, which breaks down logically complex news text into easily understandable sub-questions. The simpler input content and tasks alleviate the difficulty for LLMs in understanding the text, preventing issues such as hallucinations, where LLMs provide incorrect answers or refuse to answer, thereby enhancing the quality of task completion. Since the decomposed sub-questions require LLMs to provide intermediate answers, it can guide LLMs to integrate external knowledge and fully utilize their internal knowledge for comprehensive analysis when answering questions.
(2) Furthermore, we employ F1 score, precision, and recall to conduct a more detailed evaluation of each model's performance. It can be observed that DYNAMO achieves optimal performance across all these metrics. High precision and high recall indicate that DYNAMO identifies as many positive samples as possible with a high degree of confidence. This means that we have identified a significant number of real news, and our predictions for these real news are reliable. This is highly beneficial for our subsequent new knowledge extraction, as our new knowledge is derived from the texts and reasoning processes of real news.
(3) Compared to traditional methods, LLM-based approaches do not demonstrate significant advantages in terms of various metrics. This suggests that using LLMs does not necessarily enhance the performance of fake news detection and may even make it difficult for the models to learn the hidden relationships between data. LLMs can only fully exert their potential when appropriate input content is designed. 

\begin{table}
\renewcommand\tabcolsep{6 pt}
\centering
\begin{tabular}{ccccc}
\toprule
\multirow{2}{*}{Methods} & \multicolumn{2}{c}{\textbf{Hover}} & \multicolumn{2}{c}{\textbf{Feverous}} \\
\cline{2-3} \cline{4-5} 
 & \textbf{w/o kg} & \textbf{w/ kg} & \textbf{w/o kg} & \textbf{w/ kg}  \\
 
\midrule

LLaMA & 0.5000	& 0.5167   & 0.6180	& 0.6342\\
Flan-T5 &0.5707  & 0.5807 	& 0.5667 	& 0.5753\\
Phi3-mini & 0.5790 & 0.5850  & 0.5990 & 0.6090 \\
\midrule

Ours + LLaMA & 0.5241	& 0.5441	 & 0.6233 	& 0.6550\\
Ours + Flan-T5   & 0.5962 & 0.6077 &0.5874	&0.5924 \\
Ours + Phi3-mini  & 0.6007	 & 0.6250  & 0.6042	 & 0.6167\\
\bottomrule
\end{tabular}
\caption{Performance comparison across different LLM backbones with and without the knowledge graph. The evaluation metric is accuracy in the table.}
\label{tab:llm}
\end{table}

\subsection{Evaluation on Knowledge Graph Update}
\label{sec:kgrenew}

Table~\ref{tab:gainhover} and Table~\ref{tab:gainfever} demonstrate the improvement in detection accuracy brought about by updates to the knowledge graph. To verify the contribution of these updates, we divide each of the two datasets into three subsets. These subsets contain news reports on the same news events but with different focuses and labels, allowing us to validate whether newly acquired knowledge from previously encountered similar news events can assist in the detection of unseen news. 
To better validate the model's ability to handle unseen news, we reduce the external knowledge provided to the model regarding the last two data subsets of Hover and Feverous. Specifically, the initial knowledge graph used in this experiment is constructed solely based on the relevant factual texts from the 1st subset of Hover and Feverous, excluding any information from other subsets, and subsequent updates are made on this foundation.
In the table header, ‘‘Hover2+kg1" indicates that when detecting the 2nd subset of Hover, the knowledge graph updated during the detection of the 1st subset is used as external knowledge to assist in the detection process. As shown in Table~\ref{tab:gainhover} and Table~\ref{tab:gainfever}, through updates to the knowledge graph, the detection accuracy has been improved by an average of 1\%.
This is because our model extracts reliable and concise new knowledge from real news, filling the knowledge gap in LLMs and knowledge graphs regarding newly occurring news. The new knowledge enables LLMs to gain some understanding of new news events when they are unable to update their internal knowledge, thereby making more accurate judgments on the authenticity of the news.

\begin{figure}[]
\centering  
\subfigure{
\includegraphics[scale=0.13]{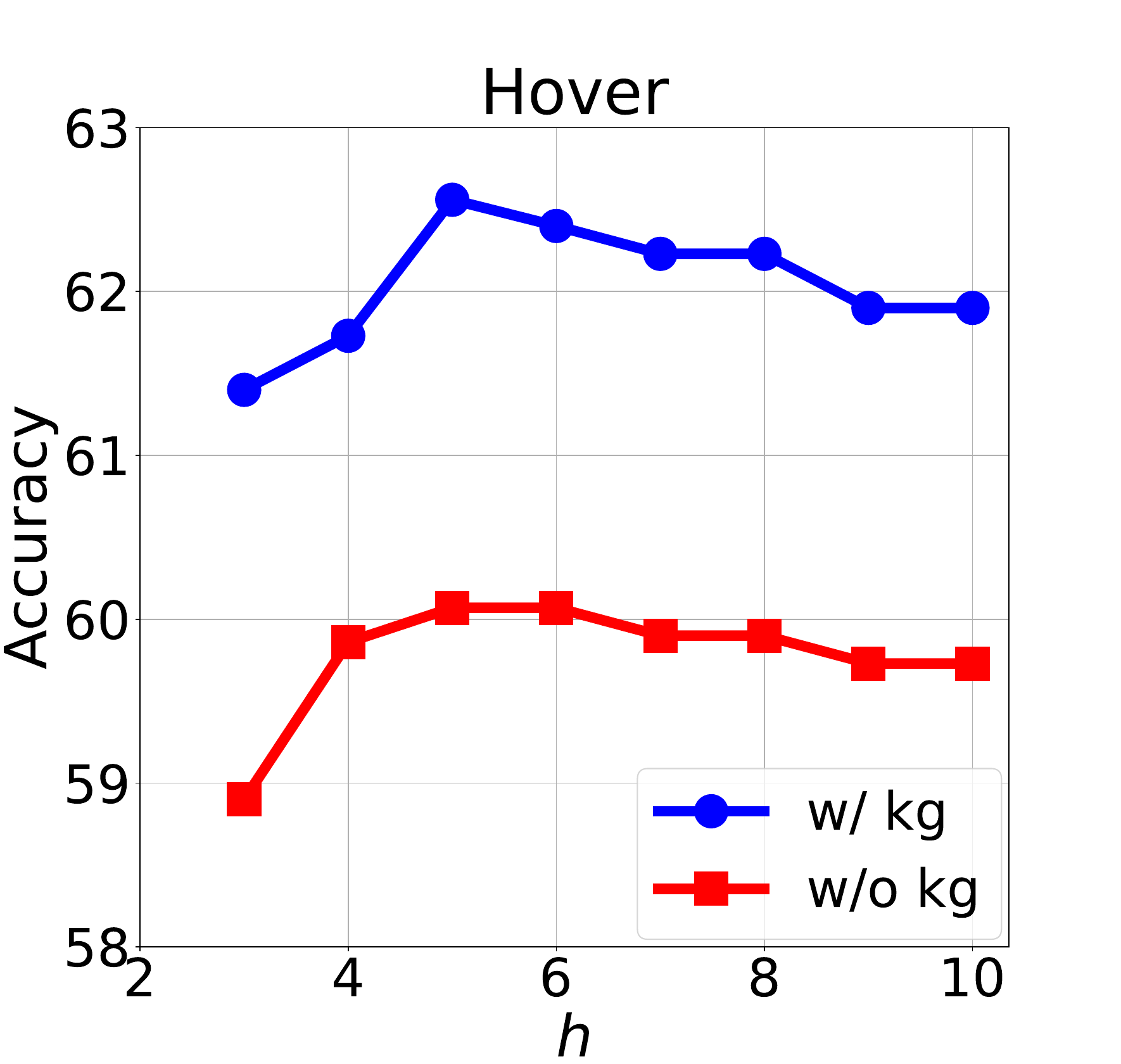}  
}
\subfigure{
\includegraphics[scale=0.13]{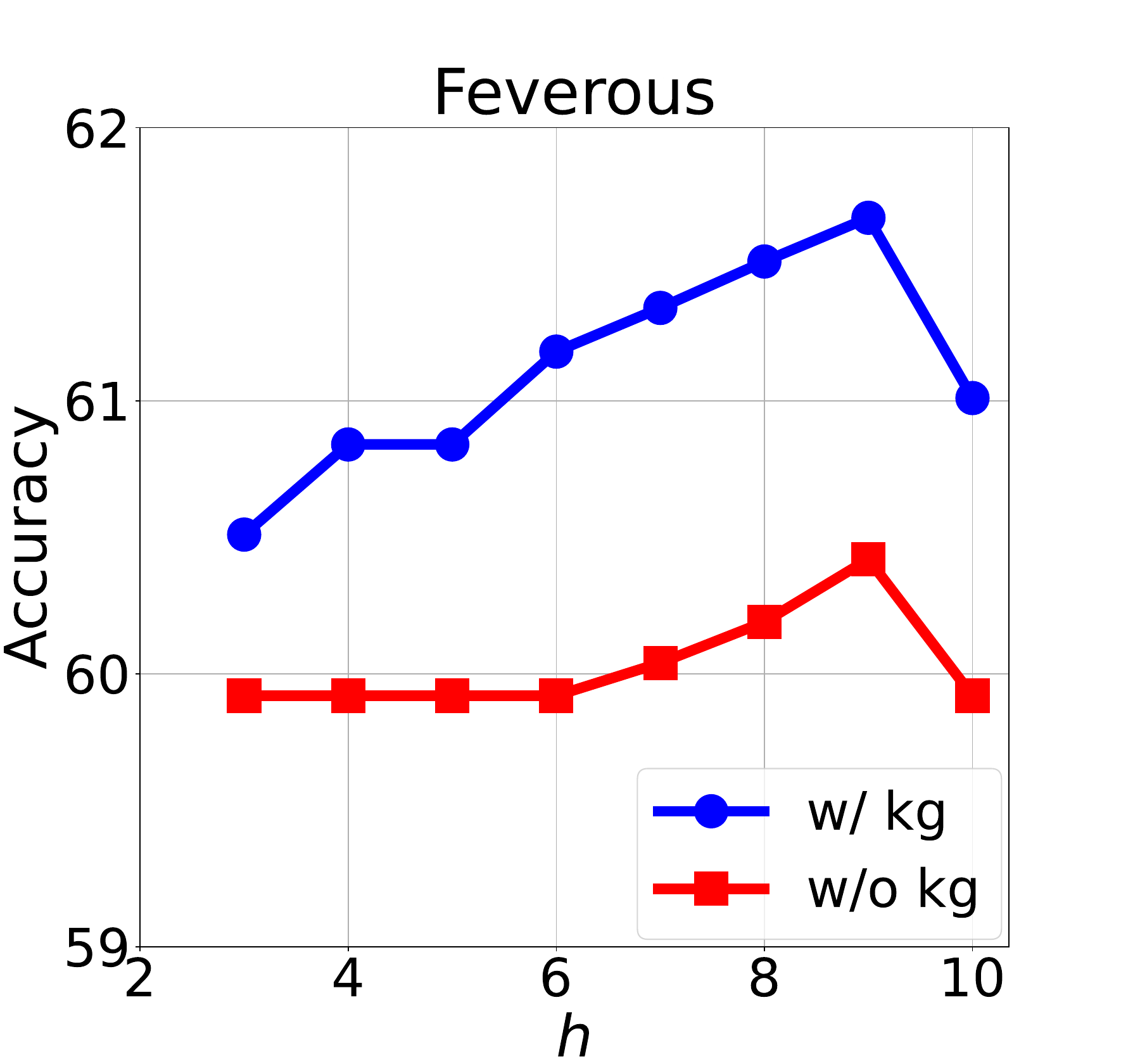}
}
\caption{Evaluation of the accuracy by varying the height limit $h$ of the Monte Carlo Tree Search.}   
\label{fig:Abla}    
\end{figure}

\begin{figure*}
    \centering
    \includegraphics[width=16.65 cm]{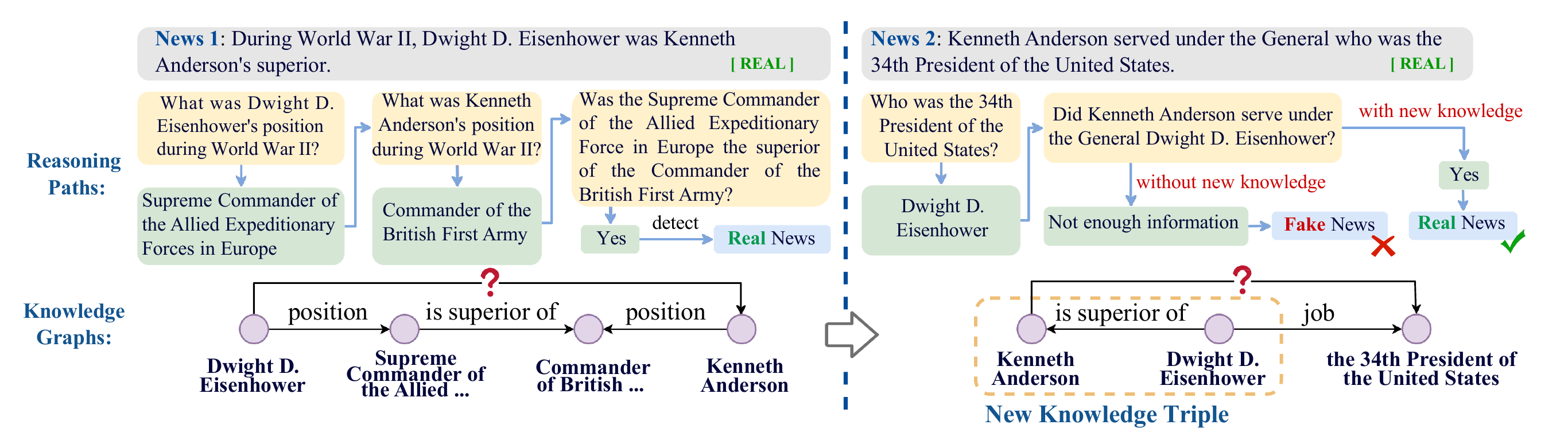}
    \caption{A set of news cases from Hover dataset. The newly acquired knowledge assists in the detection of similar news.}
    \label{fig:Case}
    \centering
\end{figure*}

\subsection{Ablation Study}
\label{sec:ablation}
To validate the impact of different modules and the parameter on the overall performance of our proposed model, we conduct multiple ablation experiments.

\textbf{LLM Backbones.} 
To demonstrate the universality of our method across different LLMs, we additionally select two LLMs: Flan-T5-XL~\cite{Scaling2024Chung} and LLaMA-13B~\cite{LLaMA2023Touvron}, as the backbone LLMs. The ‘‘w/o kg" stands for ‘‘without knowledge graph". The ‘‘w/ kg" stands for ‘‘with knowledge graph", representing the scenario where relevant knowledge triples are retrieved from the knowledge graph to serve as external knowledge. As shown in Table~\ref{tab:llm}, by comparing the performance when only the LLM is used as the detector versus when our proposed framework is applied under the same LLM settings, it can be observed that our approach enhances the detection performance of various LLMs. This is because news segmentation alleviates the difficulty for LLMs in understanding complex logic, and Monte Carlo Tree Search optimizes the segmentation process by quickly identifying suitable segmentation methods within a large search space. This experiment confirms the effectiveness of our method across various LLMs.

\textbf{External Knowledge.} 
To validate the effectiveness of the knowledge graph extracted in Section~\ref{sec:KGC}, we conduct experiments with different LLM backbones and datasets under two settings: without the involvement of the knowledge graph and with the involvement of the knowledge graph. The ‘‘w/o kg" stands for ‘‘without knowledge graph" and the ‘‘w/ kg" represents the scenario where relevant knowledge triples are retrieved from the knowledge graph to serve as external knowledge. As shown in Table~\ref{tab:llm}, the results confirm that the knowledge graph can provide additional knowledge for fake news detection, thereby improving detection accuracy.

\textbf{Hyper-parameter Sensitivity Analysis.} 
To enhance reasoning efficiency, we impose a height limit on the Monte Carlo Tree Search. To investigate the impact of the parameter height limit $h$ on detection results, we compare the accuracy of fake news detection under different height limits using the same experimental settings. Figure~\ref{fig:Abla} illustrates that, on the whole, our model demonstrates robustness to the parameter $h$, and variations in $h$ do not cause abrupt changes in detection accuracy. For the Hover dataset, the model performs best when $h$ is set to 5, regardless of whether the knowledge graph is involved. In contrast, for the Feverous dataset, the optimal performance of the model occurs when $h$ is set to 9. This is because compared to Hover, news in Feverous tend to be more verbose, containing more complex logical relationships. If $h$ is set too low, the reasoning process for some samples may not be completed. Conversely, for the Hover dataset, an excessively high $h$ can lead to excessive deep reasoning, potentially introducing too much interfering information and affecting the judgment of the LLM. Therefore, we set $h$ to 5 and 9 respectively for the Hover and Feverous datasets.

\subsection{Case Study}
\label{sec:casestudy}

To visually demonstrate the enhancement effect of dynamically updated knowledge on fake news detection, we select a set of news cases from the Hover dataset and present the entire process of their detection and the process of updating the knowledge graph. As shown in Figure~\ref{fig:Case}, News 1 and News 2 constitute a pair of news that are content-related but differ in their wording and focus. 

During the initial stage of detecting News 1, the model lacks relevant information about the relationship between Dwight D. Eisenhower and Kenneth Anderson, and therefore can not directly provide a detection result. During the Monte Carlo Tree Search process, the LLM leverages its internal knowledge and the initial knowledge graph to obtain the positions held by Dwight D. Eisenhower and Kenneth Anderson during World War II, and recognizes the hierarchical relationship between the two positions. Based on this, the model integrates the information contained in the reasoning path, deduces the relationship between the two individuals, and judges the news to be true. Subsequently, the model extracts ‘‘Dwight D. Eisenhower was Kenneth Anderson's superior" from the news content as new knowledge and updates it into the knowledge graph. For the detection of News 2, if there is no additional new knowledge, the model still would not know the relationship between the two individuals and cannot guide itself to think step-by-step by splitting the news. Therefore, in situations where there is insufficient information to support a judgment, the model tends to deem the news as fake. However, with the incorporation of new knowledge, the model can directly retrieve the relevant triples of knowledge to obtain the relationship between the two individuals, thereby swiftly judging the news as true. 

Thus, using Monte Carlo Tree Search to split the news can guide the model to think step-by-step and integrate the information, thereby obtaining new relationships between entities. Through knowledge updating, the model can possess more adequate knowledge to judge similar news.

\section{Conclusion}

In this work, we achieve dynamic knowledge updating in fake news detection, , thus providing a feasible approach to address the timeliness requirements in this field. Specifically, we leverage knowledge graphs to enable continuous updating of new knowledge and integrate large language models to perform the dual functions of news authenticity detection and verification of the correctness of new knowledge, thereby solving two key issues: ensuring the authenticity of new knowledge and deeply mining news semantics.

Since fake news often contains a mixture of true and false information, and useful real information may be embedded within false news, in the future, we plan to implement fine-grained acquisition of new knowledge. For news detected as fake, we will distinguish between the true and false parts and extract new knowledge from the true parts, thereby maximizing the utilization of the detected news.

\section*{Acknowledgements}
This work was supported by the National Natural Science Foundation of China (No. 62302333, 92370111, 62422210, 62272340, 62276187), and Hebei Natural Science Foundation (No. F2024202047) .

\bibliographystyle{named}
\bibliography{ijcai25}

\end{document}